\documentclass[10pt, conference]{IEEEtran}

\usepackage{graphicx}
\usepackage{subcaption}
\usepackage[cmex10]{amsmath}
\usepackage{amsthm}
\usepackage{comment}

\usepackage{hyperref}
\usepackage[latin1]{inputenc}
\usepackage{color}
\usepackage[normalem]{ulem}
\usepackage{amssymb}
\usepackage{fancyhdr}
\begin{document}
\title{BoWFire: Detection of Fire in Still Images \\by Integrating Pixel Color and Texture Analysis}

%------------------------------------------------------------------------- 
% change the % on next lines to produce the final camera-ready version 
\newif\iffinal
%\finalfalse
\finaltrue
\newcommand{\jemsid}{14}
%------------------------------------------------------------------------- 


\iffinal
  \author{%
    \IEEEauthorblockN{Daniel Y. T. Chino, Letricia P. S. Avalhais, Jose F. Rodrigues Jr., Agma J. M. Traina}
    \IEEEauthorblockA{%
      Institute of Mathematics and Computer Science, University of Sao Paulo\\
      Sao Carlos, Brazil\\
     % line 4: Web page: \href{http://www.institution1.eud/~firstauthor}{www.institution1.eud/$\sim$\{john,joao\}}}
      \{chinodyt,letricia,junio,agma\}@icmc.usp.br}
  %\and
  %  \IEEEauthorblockN{Authors Name/s per 2nd Affiliation}
  %  \IEEEauthorblockA{%
  %    line 1 (of Affiliation): dept. name of organization\\
  %    line 2: name of organization, acronyms acceptable\\
  %    line 3: City, Country\\
  %    line 4: Email:  \href{mailto:name@xyz.com}{name@xyz.com}}
  }
\else
  \author{Sibgrapi paper ID: \jemsid \\ }
\fi

\maketitle
\begin{abstract}
Emergency events involving fire are potentially harmful, demanding a fast and precise decision making.
The use of crowdsourcing image and videos on crisis management systems can aid in these situations by providing more information than verbal/textual descriptions.
Due to the usual high volume of data, automatic solutions need to discard non-relevant content without losing relevant information.
There are several methods for fire detection on video using color-based models.
However, they are not adequate for still image processing, because they can suffer on high false-positive results.
These methods also suffer from parameters with little physical meaning, which makes fine tuning a difficult task.
In this context, we propose a novel fire detection method for still images that uses classification based on color features combined with texture classification on superpixel regions.
Our method uses a reduced number of parameters if compared to previous works, easing the process of fine tuning the method.
Results show the effectiveness of our method of reducing false-positives while its precision remains compatible with the state-of-the-art methods.
\end{abstract}

\begin{IEEEkeywords}
fire detection;
still images;
pixel-color classification;  
texture feature
\end{IEEEkeywords}

\IEEEpeerreviewmaketitle

\section{Introduction}

% ---------------- Motivation ------------------

Emergency situations can cause economic losses, environmental disasters or serious damage to human life. 
In particular, accidents involving fire and explosion, have attracted interest to the development of automatic fire detection systems. 
Existing solutions are based on ultraviolet and infrared sensors, and usually explore the chemical properties of fire and smoke in particle samplings~\cite{Chen2004}. 
However, the main constraint of these solutions is that sensors must be set near to the fire source, which brings complexity and cost of installation and maintenance, especially in large open areas.

% RESCUER
Alternative to sensors, cameras can provide visual information of wider spaces, and have been increasingly embedded in a variety of portable devices such as smartphones. 
To take advantage of this, the {\it RESCUER}\footnote{Project  FP7-ICT-2013-EU-Brazil  - "RESCUER - Reliable and Smart Crowdsourcing Solution for Emergency and Crisis Management"} Project is developing an emergency system to support Crisis Control Committees (CCC) during a crisis situation.
%To take advantage of this, the \textit{X}\footnote{Hidden due to double blind review.} Project is developing an emergency system to support Crisis Control Committees (CCC) during a crisis situation.
%The system allows witnesses, victims or rescue staff present at an emergency location to send images and videos of the incident to a crowdsourcing mobile framework.
%Considering a disaster scenario on crowded event or a densely populated area, a huge amount of data arriving from the same event may be difficult to the specialists to analyze them. 
%For this reason, a data analysis solution is being investigated with the goal to identify the interesting information among all data by discarding what may not be related to the incident.
%With these previous automatic analysis, the decisions making actions and future prevention strategies can be more effective since they will be based on more precise information.
The system under development in the \textit{RESCUER} Project allows witnesses, victims or rescue staff, present at the emergency location, to send images and videos of the incident to a crowdsourcing mobile framework.
However, this approach might lead to a volume of information flowing at a rate higher than what human specialists are able to analyze.
For this reason, as part of the \textit{RESCUER} Project, an automated data analysis solution is under investigation aimed at identifying the most relevant pieces of information, filtering out irrelevant data.
Relevance here refers to images that actually contain fire elements, and that can effectively assist in the decision making of a given crisis.

%\sout{Images and videos acquired during crises are processed by computer vision-based techniques, aiming at modeling or detecting patterns and behaviors that may contribute to ``understand'' their content.}
%In fact, computer vision techniques have been explored to automatic detection of fire on videos.
Several methods regarding to fire detection on videos have been proposed in the last years.
These methods use two steps to detect fire.
First, they explore the visual features extracted from the video frames (images); second, they take advantage of the motion and other temporal features of the videos~\nocite{Qiu2012}\cite{Kim2014}.
In the first step, the general approach is to create a mathematical/rule-based model, defining a sub-space on the color space that represents all the fire-colored pixels in the image.
There are several empirical models using different color spaces as RGB~\cite{Chen2004}, YCbCr~\cite{Celik2009}, CIE Lab~\cite{Ha2012} and HSV~\cite{Zhao2011}.
%There are several empirical models using different color spaces as RGB~\cite{Chen2004}\cite{Kim2014}, YCbCr~\cite{Celik2009}\cite{Shon2014}, CIE Lab~\cite{Wang2009}\cite{Ha2012} and HSV~\cite{Yamagishi1999}\cite{Zhao2011}.
In these cases, the limitation is the lack of correspondence of these models to fire properties beyond color. 
%In these cases, the main limitation is the lack of correspondence of these models with other fire properties, since only the color is used, without taking information of regions or the relation between the pixels and their neighborhood.
The problem is that high illumination value or reddish-yellowish objects lead to a higher false-positive rate. 
These false-positives are usually eliminated on the second step through temporal analysis.

In contrast to such methods, our proposal is to detect fire in still images, without any further (temporal) information, using only visual features extracted from the images.
To overcome the problems aforementioned, we propose a new method to detect fire in still images that is based on the combination of two approaches: pixel-color classification and texture classification. 
%The use of texture feature is based on the observation that fire presents particular textures that are often distinct from other regions that are not fire but are fire-like colored.
The use of color is a traditional approach to the problem; whilst, the use of texture is promising, because fire traces present particular textures that permit to distinguish between actual fire and fire-like regions. %\cite{Rudz2013} 
We show that, even with just the information present in the images, it is possible to achieve a high accuracy level in such detection. 

%It is important to note that these methods were developed for videos and usually explore the advantage of motion and other temporal features combined with visual features.
%In spite of related methods were developed for videos, what concerns to the image level, most of the proposals create a color model used to define a sub-space on the color space in which all fire color pixels should be represented.  

The main contribution of this research is the proposal of \textit{BoWFire} (\textit{Best of both Worlds Fire detection}), a novel method to detect fire in still images.
By merging color and texture information, our method showed to be effective in detecting true-positive regions of fire in real-scenario images, while discarding a considerable quantity of false-positives.
%Also, no parameters were added to the method, depending only on the parameters of the techniques used, which are more intuitive to tune than the parameters presented in related works.
Our method uses fewer parameters than former works, what leads to a more intuitive process of fine tuning the automated detection.
Regarding these claims, in the experiments, we systematically compare \textit{BoWFire} with four works that currently define the state-of-the-art, that is, the works of Celik {\it et al.} \cite{Celik2009}, Chen {\it et al.} \cite{Chen2004}, Rossi {\it et al.} \cite{Rossi2011}, and Rudz {\it et al.} \cite{Rudz2013}.
The remaining of this manuscript is organized as follows: Section~\ref{sec:background} briefly surveys previous approaches related to fire detection methods; 
Section~\ref{sec:proposal} presents the proposed method using color and texture segmentation; 
Section~\ref{sec:experiments} describes the experimentation; 
Section~\ref{sec:results} presents the discussion about the results and, finally, Section~\ref{sec:conclusion} concludes this study.

% Computer vision based method for real-time fire and flame detection
% A novel algorithm for fire/smoke detection based on compter-vision
% Fire surveillance method based on quaternionic wavelet features
% Discrete cosine coefficients as images features for fire detection based on computer vision
% Flame recognition algorithm research under complex backgroud (2010) Qing Liu

\section{Related Works}
\label{sec:background}

%fire detection on videos
%adding motion and temporal data.
%However there are 
%
%There are an extensive literature on forest and urban fire detection on videos \cite{Chen2004,Celik2009, Rudz2009, Rossi2011,Rudz2013}.
%
% different color spaces
%YCbCr Celik2009, YUV Rossi2011, HSI and RGB Chen2004
%adn YCbCr has a better discrimination regarding fire pixels \cite{Celik2009, Rudz2009}.

A fire detection method based on rules was proposed in the work of Chen \textit{et al.}~\cite{Chen2004}.
They define a set of three rules using a combination of the RGB and the HSI color spaces; the user, in turn, must set two threshold parameters to detect fire pixels.
%To dismiss false-positives the authors proposed a dynamic analysis based on the fire motion.
%Since we are interested only on fire detection on still images, we will focus our analysis only on the image parts of the method.
Another method based on color was proposed by Celik \textit{et al.}~\cite{Celik2009}, who conducted a wide-ranging study regarding the color of fire pixels to define a model.
%This method defines a set of five mathematical rules based on the YCbCr color space; this was because the YCbCr has a better discrimination regarding fire~\cite{Celik2009}\cite{Rudz2013}.
%These rules compare the intensity of the YCbCr channels and the user must define a threshold parameter.
This method defines a set of five mathematical rules to compare the intensity of the channels in the YCbCr color space; this was because the YCbCr has a better discrimination regarding fire~\cite{Celik2009}\cite{Rudz2013}.
Also in this work, the user must set a threshold for one of the rules. 

Rossi \textit{et al.}~\cite{Rossi2011} proposed a method to extract geometric fire characteristics using stereoscope videos.
One of the steps is a segmentation based on a clustering algorithm, in which the image is divided into two clusters based on the channel V of the YUV color space.
The cluster with the highest value of V corresponds to fire.
Thereafter, Rossi \textit{et al.} used a 3D-Gaussian model to classify pixels as fire.
In this method, the accuracy of the classification depends on a parameter provided by the user. 
This method presents limitations, since the authors assume that the fire is registered in a controlled environment.

Rudz \textit{et al.}~\cite{Rudz2013} proposed another method based on clustering.
%Instead of using the YUV color space, Rudz \textit{et al.} computes four clusters using the blue chrominance Cb of the YCbCr color space.
Instead of using the YUV color space, Rudz \textit{et al.} computes four clusters using the channel Cb of the YCbCr color space.
The cluster with the lowest value of Cb refers to a fire region.
A second step eliminates false-positive pixels using a reference dataset.
The method treats small and large regions with different approaches; small regions are compared with the mean value of a reference region, while large regions are compared to the reference histogram.
This comparison is made for each RGB color channel.
%Although a post-processing step eliminates a high rate of false-positives, the user must define three constants for the small regions, and three thresholds for the large regions, resulting in a total of six parameters.
The user must set three constants for the small regions, and three thresholds for the large regions, resulting in a total of six parameters.

In comparison to the previous works, our method improves the state-of-the-art by using texture, beyond color, to reduce false-positives; and by using a smaller set of parameters, an important characteristic to fine tune the detection process. 
Besides, the parameters of former methods do not carry physical significance, therefore, they are less intuitive to adjust.

%Since these aforementioned methods are color-based, they suffer with the high rate of false-positives.
%Also, these methods use too many parameters and are very sensitive in respect to their tuning. 
%Another problem of their parameters is the lack of physical significance.
%The majority of these parameters are thresholds based on the color intensity values or multiple constants determined empirically, making the fine tuning of their methods very troublesome.

% edge based methods
%An edge-detection of flames for images was proposed in~\cite{Qiu2012}. 
%The main steps of the algorithm consist on use a Gaussian filter to eliminate noise; apply Sobel operator for finding basic edges and adjust some autoadaptive parameters to improve the edges.
%Another algorithm based on edge-detection of flames was developed for forest fire detection~\cite{Zhang2008} in videos. 
%First a Laplacian operator is applyed in a region based on the red color channel and after a eight-connected boundary chain code is used to extract the fire contour. 
%After this, they use the temporal wavelet to analyse the variance of consecutive Fourier descriptors. 
%The drawback of these methods is that the flames edges are detected in images that contains only one flame with a strong luminance suggesting high contrast with the background.

\section{Basic Concepts and Notations}

In this section we present important concepts related to the problem under analysis.
An image can be defined as a set of pixels $I = \{P_i | 0 \leq i < n\}$, where $n$ is the total number of pixels in the image. 
Each pixel is a tuple $P_i = (R_{i}, G_{i}, B_{i})$, where the values of $R_{i}, G_{i}$ and $B_{i}$ represents the intensity of each channel of the RGB color space.
Global or local information can be extracted from images.
A feature extraction is a function $F$ that  for a given image $I$, generates a feature vector $V \in \mathbb{R}^d$, of size $d$.

Let a set $\mathcal{T} \subset \mathbb{R}^d$ of tuples of size $d$ and a set of possible labels $\mathcal{L}$.
Given a training set in which every $t_{i} \in \mathcal{T}$ has a label $c_{i}$ assigned by an expert, where $c_{i} \in \mathcal{L}$; a supervised classifier $C$ must build a model capable to predict the label of a new data item.
Given a tuple $x \in \mathbb{R}^d$, a classifier can be defined as the function $C(x) = c$, where $c \in \mathcal{L}$.
Among the most used classifiers are the Naïve-Bayes~\cite{Garcia2013} and $K$-Nearest Neighbors (\textit{KNN})~\cite{Duda2000}.
We refer to the above definitions in the rest of this work. 
%Let $s = (s_{0}, \dots, s_{n-1})$ be a tuple of size $n$ and $s_{i} \in \mathbb{R}$, $0 \leq i < n\}$.
%Let $\mathcal{S}_{n}$ be the set of all $s$.

\subsection{Feature Extraction}

%One possible way to analyze images is by using textual information that describes each image.
%This type of approach uses visual perception and in general requires a supervised task.
%Considering that the amount of data is constantly growing, this approach becomes impracticable.
%Also, the image may have some details that are hard to describe using textual information \cite{Zhang2002}.
%Another approach automatically describes the image using only the contents of the image (image descriptor).
%An image descriptor is a pair of a feature description extracted from the image and a similarity measure between the extracted features.
%\sout{In many computational applications,} 
Images are processed by means of extracted features.
The features extracted from a given image correspond to numerical measurements that describe visual properties.
Such properties are able to discover and represent connections between pixels of the whole image (global)~\cite{Torres2006}, or of small regions of the image (local)~\cite{Shaban2013}.
Low-level descriptors~\cite{Deselaers2008}, as those base on color, shape and texture, are frequently used. 
%An image can be described using low level descriptors~\cite{Deselaers2008} which give a description about color, texture and shape.

Usually, color-feature extraction methods have a low computing cost.
%\sout{In the computation of the color histogram, for example, it is possible to compute a second metric -- the color moments -- without significantly compromising the performance~\cite{Datta2008}.}
%The color moments describe the probability distribution of the image colors, which is useful for many computations \cite{Datta2008}.
Color Layout~\cite{Jalab2011} is an example of color extraction, it describes the space distribution of the colors. 
The color histogram also allows to compute the color moments~\cite{Datta2008} to describe the probability distribution of the image colors.
%The color histogram also allows to compute the color moments~\cite{Datta2008} to describe the probability distribution of the image colors.
%Another color feature extractor is the Color Layout~\cite{Jalab2011}, which describes the space distribution of the colors.
%Besides color, texture is a common feature in image processing.
Shape information, in turn, is considered the closest approximation to the human perception of an object's image ~\cite{Yang2008}.
Feature extractors of this depend on a pre-processing step that segments and detects the border of the objects.
There are various methods to extract shape features, as the Zernike moments~\cite{Hosny2008} and the Fourier descriptors~\cite{Chen2008}. % and the contour salience descriptors~\cite{Torres2007}.
Texture is also a common feature in image processing. 
It is important because, together with color, it describes the surface of naturally-occurring phenomena, as fire, smoke, and water.
%Texture is a common feature on images, since it is present in many daily objects, as fire, smoke and water.
%\sout{These objects' textures can be described, for example, by the roughness and homogeneity of their surfaces~\cite{Saipullah2012}.} 
Classical feature extractors of texture are LBP~\cite{Ojala1994} and Haralick~\cite{Haralick1973}.
%\cite{Aksoy2001}

%\subsection{Local Binary Patterns}
%\label{sec:LBP} 

The Local Binary Pattern (LBP) is a texture feature extractor that considers the neighborhood of a pixel~\cite{Ojala1994}.
The LBP can be used in several applications, as helmet detection on motorcyclists~\cite{Silva2014}, MR image retrieval systems~\cite{Varghese2014}, and image segmentation~\cite{Vargas2014}.
The LPB can be performed in gray scale, in a region of 3$\times$3 pixels. 
For each pixel $P_{i}$ from the neighborhood of the central pixel $P_{c}$, a binary code is created by assigning the value $1$ if $P_{i} > P_{c}$, or $0$, otherwise.
Then, the histogram of codes is used as a feature vector.

In order to make the LBP rotation invariant, a variation of the original algorithm shifts the code until it reaches its minimum value~\cite{Maenpaa2003}.  
Another variation is defined as \textit{uniform patterns}~\cite{Ojala2002}. 
A code is called uniform if the binary pattern contains at most two bitwise transitions from 0 to 1 (or vice versa).
%The histogram is obtained by taking one bin for each uniform pattern and a single bin for all of the non-uniform patterns.
The histogram is obtained by taking one bin for each uniform pattern and a single bin for the non-uniform patterns.

%All these features extracted from the images can be represented as a feature vector.
%The feature vector can be interpreted as a point in a m-dimensional space \cite{Guyon2006}.
%This way, given two images and their corresponding feature vectors, it is possible to say if they are similar.
%Two images are considered similar if they have a low distance between their feature vectors.
%The most common distance functions are the ones based on the Minkowski family \cite{Wilson1997}, Euclidian, Manhattan and Chebychev.
%There are others distances functions, as the Canberra distance and Kullback-Leibler divergence.

\subsection{Superpixel Generation}
\label{sec:superpixel} 

Superpixel algorithms group pixels into atomic regions with similar homogeneity.
Doing so, superpixels can capture the image redundancy and reduce the complexity of subsequent image processing tasks.
Superpixels can be used as building blocks of many computer vision algorithms~\cite{Achanta2012}, in this paper, we will focus on image segmentation~\cite{Rauber2013}\cite{Gallo2014}. %\cite{Vargas2014}

A superpixel $Sp$ is defined as a subset of the image $I$ that contains pixels from a continuous region of the image.
A superpixel generation algorithm can be described as the process $S(I, K_{sp}) = \{Sp_{j} | 0 \leq j < K_{sp}\}$ that takes the image $I$ and returns $K_{sp}$ partitions.
A superpixel generation algorithm must have a good adherence to image boundaries and should improve the performance of the segmentation algorithm.
An empirical comparison between the state-of-the art superpixel algorithms was made by Achanta {\it et al.}~\cite{Achanta2012}.
Their results showed that the algorithm with the best overall performance was the Simple Linear Iterative Clustering (SLIC) technique.
The SLIC technique is an adaptation of K-Means for superpixel generation using a distance function based on the values of pixels using the Lab color space and their geometric position.
The user gives as parameter the number of superpixels $K_{sp}$ and their compactness $m$.
The algorithm then positions the centroids on a regular grid, avoiding seeding a superpixel on an edge pixel.
%\sout{To optimize the clustering step, SLIC computes the distance only for pixels within a limited region of the centroid.}

\section{Our Proposal}
\label{sec:proposal}

\begin{figure}[ht]
	\centering
	\includegraphics[width=0.9\columnwidth]{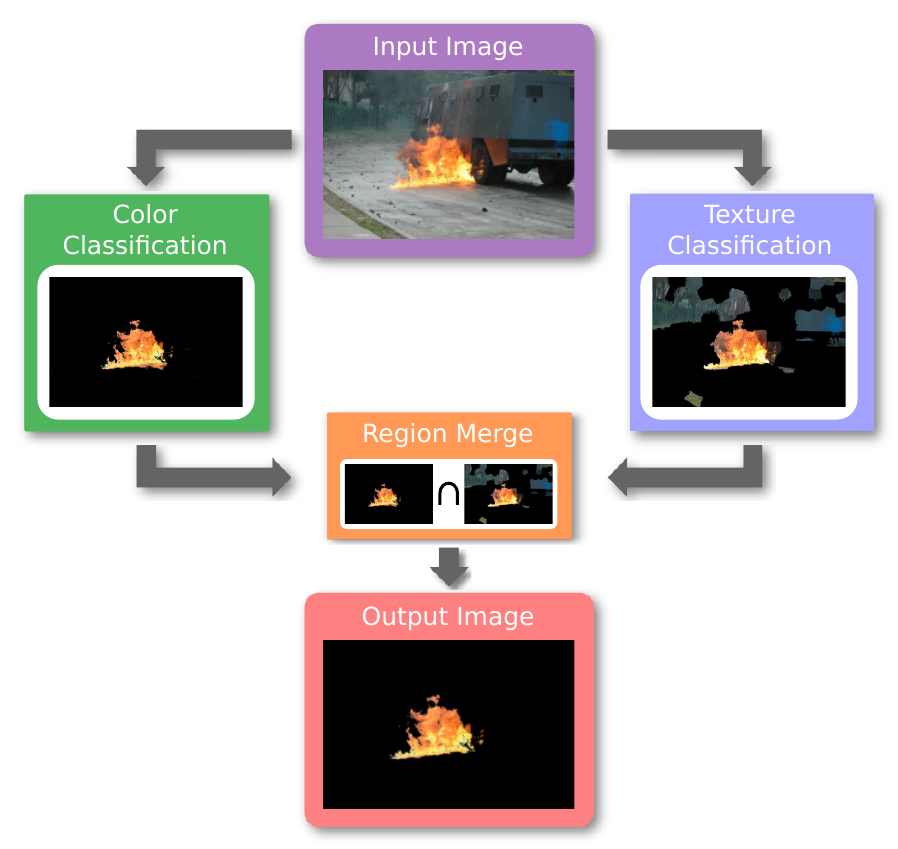}
	\caption{
		Architecture of the \textit{BoWFire} method.
	}
	\label{fig3:framework}
\end{figure}

We propose \textit{BoWFire} (Best of Both Worlds Fire detection), a novel method for fire detection in emergency-situation images. 
We explore the fact that color combined with texture can improve the detection of fire, reducing the number of false-positives as compared to related works from the literature. 
We show that such combination can distinguish actual fire from fire-like regions (reddish/yellowish) of a given image. 
The goal is to provide a more effective automated detection of fire scenes in the context of the crisis situations, as those of the \textit{RESCUER} Project. 
%As shown on the literature, color-based methods are highly capable of detecting fire on images.
%Because of the high capability of color feature to detect fire on images, the related works are, in most of the cases, color-based.
%However, they are susceptible to high false alarms while dealing with images without fire but containing fire-like color scenes, as yellow-red surfaces or sunset skylines.
%Although these scenes present similar colors to fire scenes, they have different texture aspects.
%Thus, we can add texture information to filter non-fire pixels, resulting in better results.
%On this context, the \textit{BoWFire} aims at reducing the false-positive rate detection, reducing the amount of information given to a human evaluation.
%To do so, instead of considering only information of the color of the pixel, the \textit{BoWFire} also analyses the texture aspects of the image.
%This way, we can have the best of both worlds by using color information to capture as many fire like pixels and discard false-positives with texture information.
%This way, we can combine both color and texture features of the fire.
%although good performance while dealing with fire images dataset
%pixel color classification still suffers the same problem the color models
%false-positives.
%to overcome this problem we proposed a
%added texture information to better filter false-positive pixels
%
%two branches
Figure~\ref{fig3:framework} shows the basic architecture of our proposal.
The \textit{BoWFire} method consists of three basic steps: \textit{Color Classification}, \textit{Texture Classification}, and \textit{Region Merge}.
As shown in Figure \ref{fig3:framework}, the two first steps occur in parallel to produce images in which fire-classified pixels are marked. 
Then, the output from both classifications is merged into a single output image by the \textit{Region Merge} step.

\begin{figure}[ht]
	\centering
	\includegraphics[width=0.9\columnwidth]{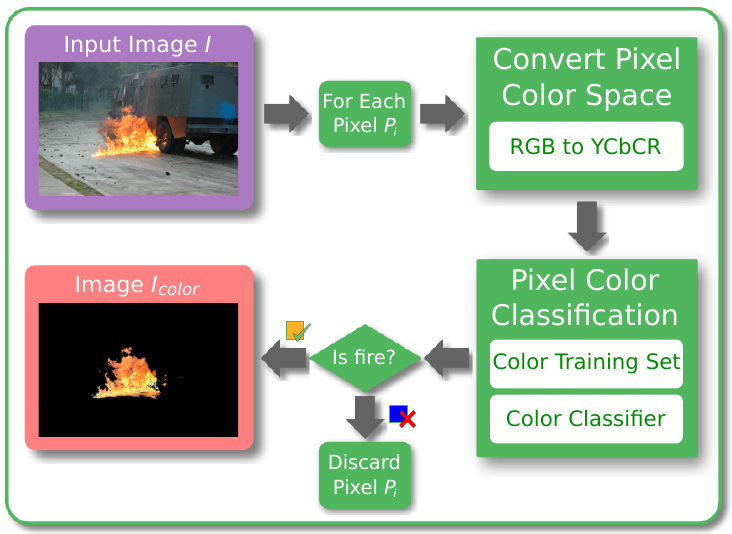}
	\caption{
		Color-based classification step.
	}
	\label{fig3:colorClass}
\end{figure}

Different from other methods, usually based on mathematical models, the use of a \textit{Color Classification} step avoids the need of a great number of parameters. 
Any machine learning classification algorithm could be used, specifically, in this work, we use Naïve-Bayes and \textit{KNN}, as detailed in Section~\ref{sec:experiments}.
%The left side of the branch is the \textit{\textit{Color Classification}} step, which distinctly from the methods in the literature that employ a mathematical modelling, we proposed a method based on classification algorithms.
By doing so, we also avoid the use of the global information of the image to classify only one pixel as opposed to other approaches; this is a desired feature because the semantics of the image may vary according to the emergency situation (small/large fire regions or day/night time). 
%By doing so, we can avoid the need of a great number of parameters.
%We can also avoid using the global information of the image to classify only one pixel, since the semantic of the image may vary according to the emergency situation (small/large fire regions or day/night time).
Figure~\ref{fig3:colorClass} presents more details of the color-based classification. 
%The \textit{\textit{Color Classification}} receives the original input image.
Given an image $I$ with $n$ pixels $P_{i}$, $0 \leq i < n$.
Each pixel $P_{i} = (R_{i}, G_{i}, B_{i})$ of the image is converted to $P'_{i} = (Y_{i}, Cb_{i}, Cr_{i})$ in the YCbCr color space, since this color space provides a better discrimination of fire regions. 
%The \textit{Color Classification} receives the original input image.
%Then each pixel $P$ of the image is converted to the YCbCr color space, since it provides a better discrimination of fire regions.
Then $P'_{i}$ goes through a \textit{Pixel-Color Classification} ($pixelClass$), which consists of a {\it Color Training Set} and a {\it Color Classifier}. 
%The YCbCr values go through a Pixel \textit{Color Classification}, which consists of a Color Training Set and a Color Classifier.
%The Pixel Color Classifier consists of a training set of correctly labeled pixels and a Naive Bayes classifier.
Then, if $pixelClass(P'_{i}) = \langle \mathrm{fire} \rangle$, $P_{i}$ is used to build the output image $I_{color}$, otherwise $P_{i}$ is discarded.
%\textcolor{red}{Our method allows the usage of different color classifiers.}
%Our architecture allows the usage of different classification algorithms.

\begin{figure}[ht]
	\centering
	\includegraphics[width=0.9\columnwidth]{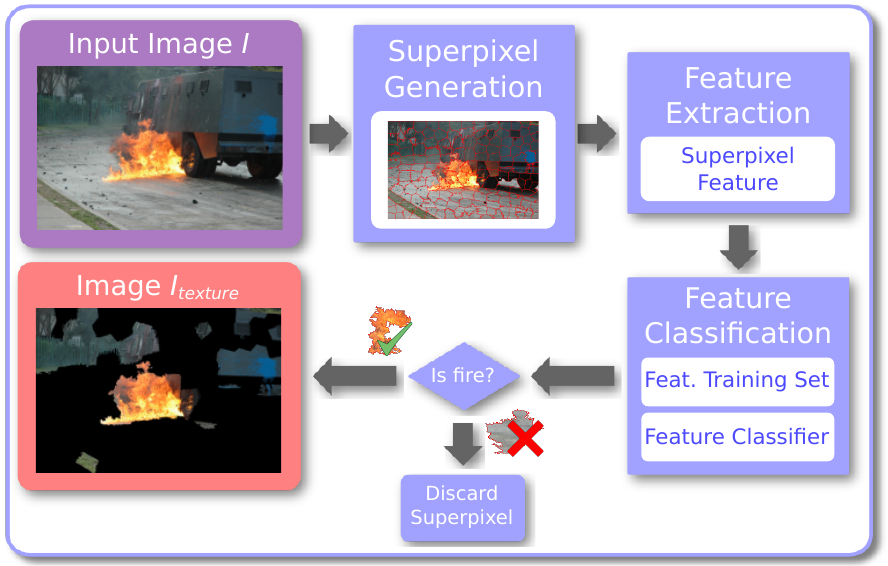}
	\caption{
		Texture-based classification step.
	}
	\label{fig3:textureClass}
\end{figure}

%The right side of the branch is the \textit{Texture Classification} step, which is responsible to add texture information.
%This step allows our proposal to discard scenes with colors similar to fire.
As mentioned earlier, the \textit{Texture Classification} step allows for a more accurate detection; however, it brings a challenge. 
%As mentioned earlier, there may be a variety of fire images according to the emergency situation.
Since there may be a variety of fire images according to the emergency situation, it is not possible to extract global features of the image because the small fire regions would vanish in the global context.
%Therefore, we could not extract global features of the image, since small fire regions would vanish on the global context of the image.
%Therefore, we extract only local features of the image. 
%In our proposal, we extract only local features of the image.
%To do so, we use superpixel methods (refer to Section \ref{sec:superpixel}) to automatically detect regions, since these methods provide regular shaped regions with similar patterns.
%To do so, we use superpixel methods to automatically detect regular shaped regions with similar patterns. %(refer to Section \ref{sec:superpixel})
Therefore, we extract only local features from regular shaped regions with similar patterns automatically detected by superpixel methods.
Figure~\ref{fig3:textureClass} presents details of the \textit{Texture Classification} step.
Given the same image $I$, we use a superpixel method $S(I, K_{sp})$ to generate a set of $K_{sp}$ superpixels $Sp_{j}$, where $0 \leq j < K_{sp}$. %, that will be used to extract local features. 
%The \textit{Texture Classification} step also receives the original input image.
%Then, we generate regions to extract local features using a superpixel algorithm. 
Next, each superpixel $Sp_{j}$ passes through a local \textit{Feature Extraction} process, resulting in a feature vector $V_{j} = (v_{j0}, \dots, v_{j(d-1)})$ of size $d$.
%Each superpixel (region) passes through a \textit{Feature Extraction} process.
Then, $V_{j}$ is classified using a \textit{Feature Classification} ($featClass$), which consists of a \textit{Feature Training Set} and a \textit{Feature Classifier}. 
%The superpixel feature is then classified using a \textit{Feature Classification} method, which consists of a \textit{Feature Training Set} and a \textit{Feature Classifier}.
If $featClass(V_{j}) = \langle \mathrm{fire} \rangle$, all pixels $P_{i} \in Sp_{j}$ are used to build the output image $I_{texture}$, otherwise they are discarded. 
%After this, the superpixel region is no longer necessary because the next step is performed in pixel-level only.
After this, the superpixel region is no longer necessary since the method is performed in pixel-level only.
%\textcolor{red}{The method allows in this step also, the use of different texture classification criteria.}
%The \textit{BoWFire} allows the use of different feature extraction algorithms, as also, different classification algorithms. 

With the outputs from the \textit{Color Classification} (image $I_{color}$), and from the \textit{Texture Classification} (image $I_{texture}$), it is still necessary to join the results in an output image $I_{classified}$.
We perform this task in the \textit{Region Merge} step.
%We perform this task in the \textit{Region Merge} step, now using only the pixel-level information.
%With the outputs from the \textit{Color Classification} and from the \textit{Texture Classification}, it is still necessary to join the results in an output image in the \textit{Region Merge} step. 
%This step intersects the images $I_{color}$ and $I_{texture}$ from each classification step. 
%The last step, the \textit{Region Merge}, is responsible for this task.
%The \textit{Region Merge} step intersects the outputs from each classification step.
According to our hypothesis, if a pixel is simultaneously classified as fire following color and texture classification, then there is a higher chance that this pixel is actual fire.
Therefore, given an image $I$ and its color and texture classifications $I_{color}$ and $I_{texture}$, the final classified image $I_{classified}$ is defined as $I_{classified}=\{P_{i}|P_{i} \in I_{color}\ \mathrm{and}\ P_{i} \in I_{texture}\}$.
That is, a given pixel is added to the final output only if it was detected in both color and texture classifications, otherwise it is discarded. 
Consequently, we dismiss false-positives from both approaches, taking advantage of the best of both worlds.

%If a pixel $P_{i} \in I_{color}$ and $P_{i} \in I_{texture}$, then $P_{i}$ is added to the final output, otherwise, $P_{i}$ is discarded.
%Thus, we can dismiss false-positives from both approaches, taking advantage of the best of both worlds.

The \textit{BoWFire} method was developed in a modularized scheme, allowing an easy way to add and set different feature extraction algorithms, as well as different classifiers.
We note that, since the \textit{BoWFire} method is fully customizable, the number of parameters is dependent only on the algorithms used in the intermediate steps. 
%The \textit{BoWFire} method was developed in a modularized scheme, allowing an easy way to add and set different feature extraction algorithms, as also, different classifiers.
%Since the \textit{BoWFire} method is fully customizable, the number of attributes is dependent only on the algorithms used in the intermediate steps.

\section{Experiments}
\label{sec:experiments}

\subsection{Configuration} 

Following, we describe the configuration of our experiments.

\begin{figure}
	\centering
	\includegraphics[]{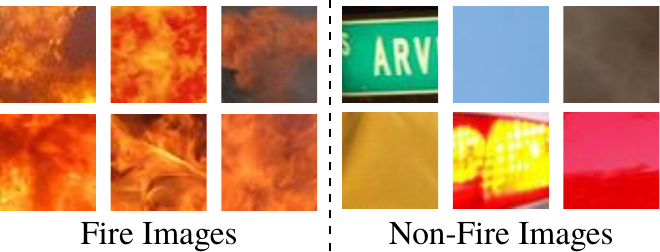}
	\caption{
		Sample images of the training dataset.
	}
	\label{fig4:trainingset}
\end{figure}

%\noindent
\textbf{Datasets:} We performed experiments using a dataset of fire images.
It consists of 226 images with various resolutions\footnote{Available at \url{http://www.gbdi.icmc.usp.br}}.
Also, it was divided in two categories: 119 images containing fire, and 107 images without fire.
The fire images consist of emergency situations with different fire incidents, as buildings on fire, industrial fire, car accidents, and riots.
These images were manually cropped by human experts. %, presenting a total of 9,681,022 fire pixels and 89,965,829 non-fire pixels.
The remaining images consist of emergency situations with no visible fire and also images with fire-like regions, such as sunsets, and red or yellow objects.
Figures~\ref{fig4:sample1}, \ref{fig4:sample2} and \ref{fig4:sample3} show some samples of this dataset.
%The images without fire presents a total of 101,404,464 non-fire pixels.
Since we are using supervised machine learning, we also used a training dataset\textsuperscript{2}. 
This second dataset consists of 240 images of 50$\times$50 pixels resolution; 80 images classified as fire, and 160 as non-fire.
Figure~\ref{fig4:trainingset} shows some samples of this dataset.
It is important to note that the non-fire images also contain red or yellow objects.
%The training dataset used on our proposed method consits of 240 images of 50$\times$50 resolution\textsuperscript{2}.
%These images were manually classified in 80 fire images and 160 non-fire images.
%This training dataset was used for both classification steps, \textit{Pixel Color Classification} and \textit{Feature Classification}.

%\noindent
\textbf{Intermediate Algorithms:} Since the goal of our method is to have a fewer number of parameters, for the \textit{BoWFire} intermediate steps, we used the following algorithms.
The \textit{Pixel-Color Classification} is done by a Naïve-Bayes classifier, using an automatic discretization method; the superpixel algorithm was the SLIC method with a modification.
Instead of using the Lab color space, we used the YCbCr space due to its discriminative property.
Since we wanted to add texture information, our implementation uses the \textit{uniform patterns} LBP.
%Other feature extraction methods were evaluated, but due to space limitations, we show only the LBP results.
Other feature extraction methods were evaluated, but were omitted due to space limitation.
The features were classified using the \textit{KNN} classification with the Manhattan Distance.

%\noindent
\textbf{Parameters:} Considering the configuration given by the choice of intermediate algorithms, the \textit{BoWFire} method needs only 3 parameters: $K_{sp}$, $m$ and $K$.
For all experiments we empirically evaluated the best values for parameters $m$ and $K$; for parameter $K$, we used the value $K=11$. %as used in \textit{KNN} classification.
Regarding to parameter $m$, we observed that a more compact superpixel generates a more regular region, which leads to a better representation of the texture feature.
In this case, the best value was $m=40$.
With these parameters, each method was executed on three different datasets: only fire images, only non-fire images, and a complete dataset with both fire and non-fire.
For each execution, we computed the confusion matrix for the classification of all pixels and calculated four measures: Precision, Recall, F1-Score, and False-Positive Rate (FPR).

\subsection{Description of the experiments}

In this section, we describe the experiments (i) on the impact of parameter $K_{sp}$, (ii) on the \textit{Color Classification} Evaluation, and (iii) on the \textit{BoWFire} Evaluation.
Next, in Section \ref{sec:results}, we report on the execution and results of these experiments.

%\noindent
\textbf{Impact of $K_{sp}$:} The first experiment evaluates the impact of the number of superpixels on the \textit{BoWFire} performance.
We vary the number of superpixels $K_{sp}$ with the following values: 50, 100, 150, 200, 250 and 300.
%Precision will determine the correctness of the method.
%Recall will tell us the capability to recover true-positives.
%F1-Score will tell us the overall... \textcolor{red}{nao sei}.
%And the False-Positive Rate will tell how much the method missed

%\noindent
\textbf{\textit{Color Classification} Evaluation:} In this experiment, we aim at evaluating the capability of the \textit{Color Classification} step proposed in this paper.
Since \textit{BoWFire} is based on a combination of two different approaches, it is important that the color-based method recovers as many fire pixels as possible.
So, Recall is the measure that closely meets this need.
Also, on this step FPR is not so important, since it will be handled on the \textit{Texture Classification} step.
%We evaluated the \textit{BoWFire} with the current state-of-the-art methods: Celik's~\cite{Celik2009}, Chen's~\cite{Chen2004}, Rossi's~\cite{Rossi2011} and Rudz's~\cite{Rudz2013}.
We evaluated the behavior of our proposed \textit{Color Classification} in comparison with the state-of-the-art, as in the works of Celik~\cite{Celik2009}, Chen~\cite{Chen2004}, Rossi~\cite{Rossi2011} and Rudz~\cite{Rudz2013}.
%\sout{We adjusted these methods according to the best parameters as indicated by their respective authors.} 
%All literature methods parameters were set with the best values according to their respective authors.
%We evaluated the behavior of our proposed \textit{Color Classification} in comparison with the state-of-the-art.

%For both these experiments, we used the same datasets described on Experiment~\ref{exp4:superpixel}.
%For each image, we also computed the overall Precision, Recall, F1-Score and FPR.

%\noindent
\textbf{\textit{BoWFire} Evaluation:} After evaluating only color, we evaluate the impact of considering texture together with color, as defined in our proposal. % methodology. 
The most important aspect of this step is to reduce false-positives without affecting the overall performance.
In this context, we analyze the \textit{BoWFire} performance, which is the combination of the \textit{Color Classification} step with the \textit{Texture Classification} step.
We also evaluated the performance of the state-of-the-art methods combined with the \textit{Texture Classification}.
We used the best value of $K_{sp}$ as obtained in the experimentation.

\section{Results and Discussion}
\label{sec:results}

%Super pixel evaluation
\begin{figure*}[t!]
	\centering
	\begin{subfigure}[b]{0.33\textwidth}
		\includegraphics[width=\textwidth]{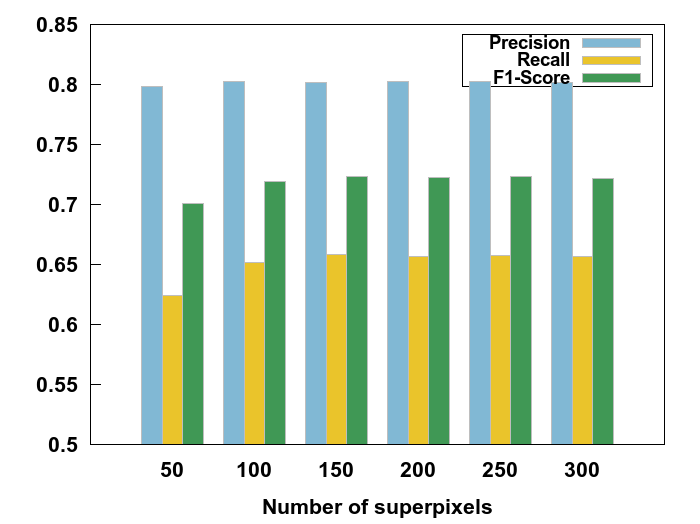}
		\caption{
			Fire dataset
		}
		\label{fig4:superPixelFire}
	\end{subfigure}
	\hfill
	\begin{subfigure}[b]{0.32\textwidth}
		\includegraphics[width=\textwidth]{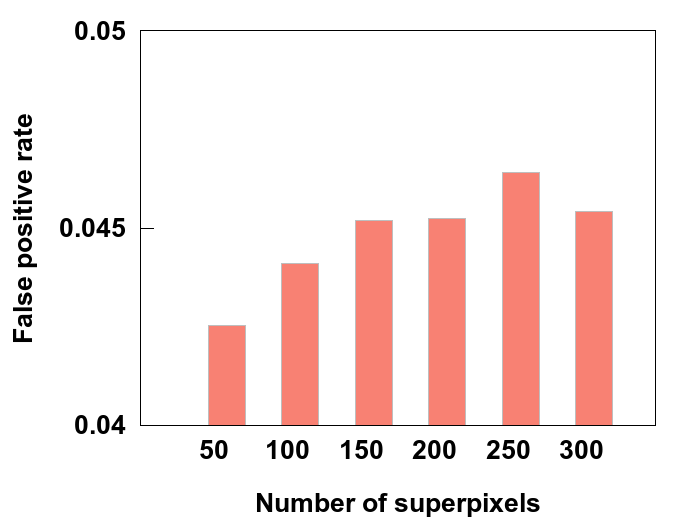}
		\caption{
			Non-fire dataset
		}
		\label{fig4:superPixelNotFire}
	\end{subfigure}
	\hfill
	\begin{subfigure}[b]{0.33\textwidth}
		\includegraphics[width=\textwidth]{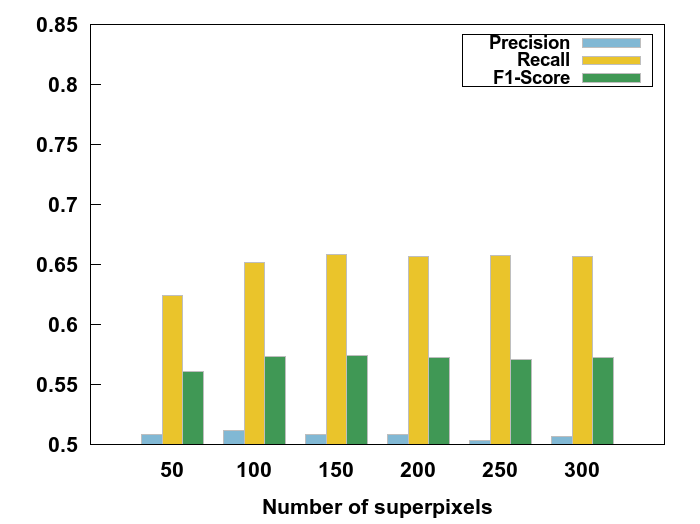}
		\caption{
			Complete dataset
		}
		\label{fig4:superPixelComplete}
	\end{subfigure}
	\caption{
		Impact evaluation of the number of superPixel $K_{sp}$ in three different datasets.
	}
	\label{fig4:superPixel}
\end{figure*}

%Color classifier comparison
\begin{figure*}[!ht]
	\centering
	\begin{subfigure}[b]{0.33\textwidth}
		\includegraphics[width=\textwidth]{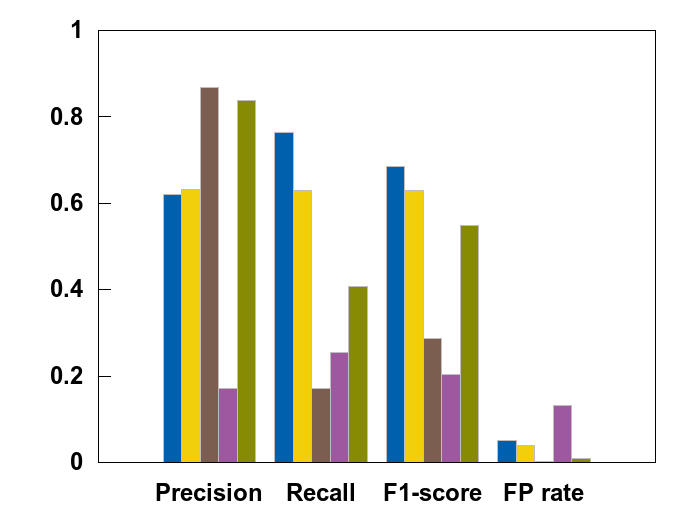}
		\caption{
			Fire dataset (only color)
		}
		\label{fig4:colorClassFire}
	\end{subfigure}
	\hfill
	\begin{subfigure}[b]{0.32\textwidth}
		\includegraphics[width=\textwidth]{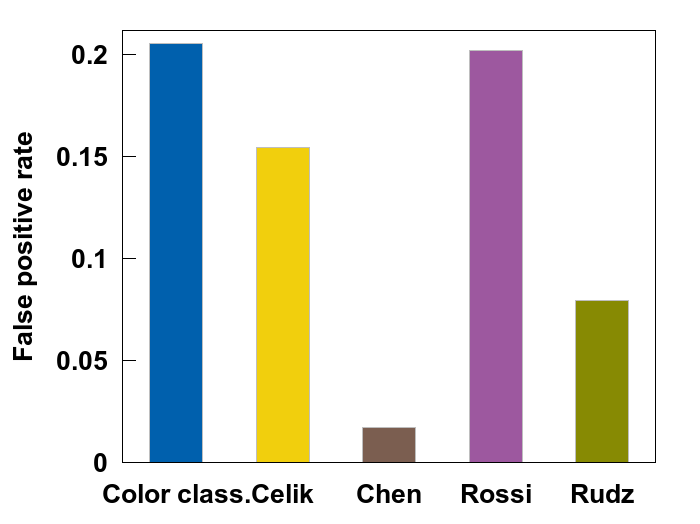}
		\caption{
			Non-fire dataset (only color)
		}
		\label{fig4:colorClassNotFire}
	\end{subfigure}
	\hfill
	\begin{subfigure}[b]{0.33\textwidth}
		\includegraphics[width=\textwidth]{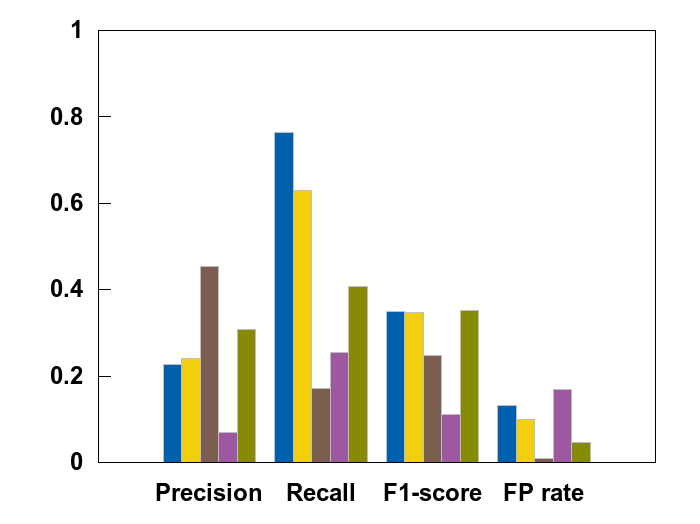}
		\caption{
			Complete dataset (only color)
		}
		\label{fig4:colorClassComplete}
	\end{subfigure}
	\\
	\begin{subfigure}[b]{0.33\textwidth}
		\includegraphics[width=\textwidth]{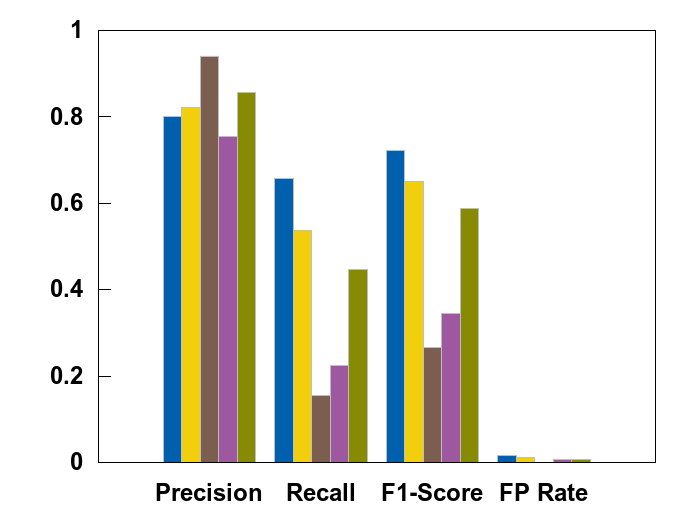}
		\caption{
			Fire dataset (adding texture)
		}
		\label{fig4:textureClassFire}
	\end{subfigure}
	\hfill
	\begin{subfigure}[b]{0.32\textwidth}
		\includegraphics[width=\textwidth]{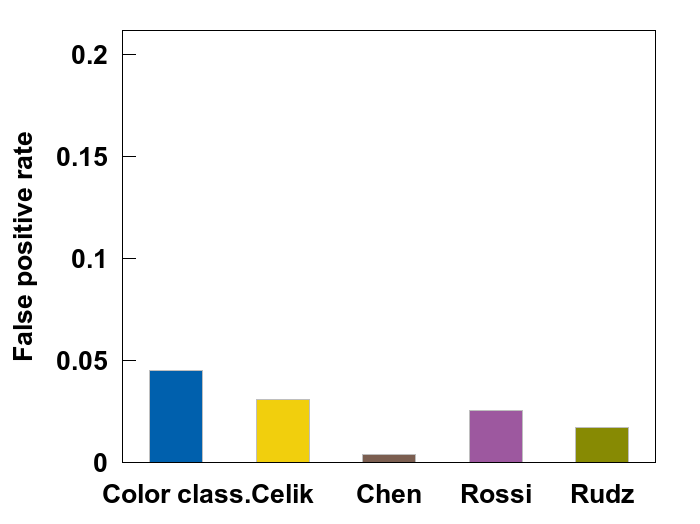}
		\caption{
			Non-fire dataset (adding texture)
		}
		\label{fig4:textureClassNotFire}
	\end{subfigure}
	\hfill
	\begin{subfigure}[b]{0.33\textwidth}
		\includegraphics[width=\textwidth]{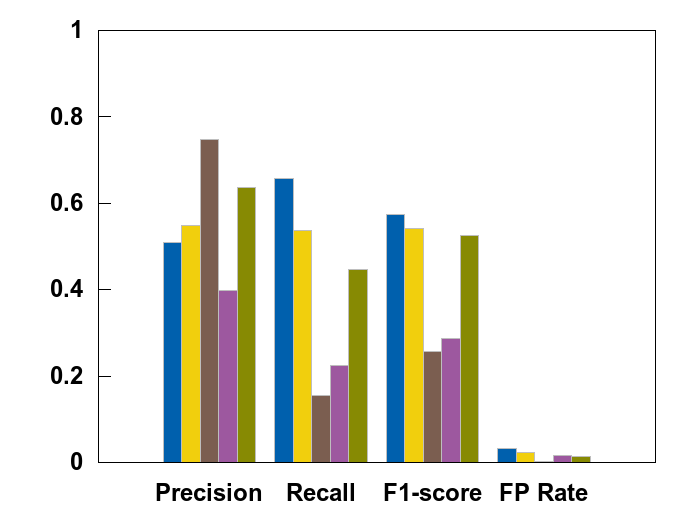}
		\caption{
			Complete dataset (adding texture)
		}
		\label{fig4:textureClassComplete}
	\end{subfigure}
	\\
	\begin{subfigure}[b]{\textwidth}
		\centering
		\includegraphics[height=0.27cm]{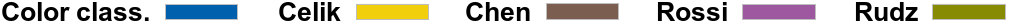}
	\end{subfigure}
	\caption{
		Evaluation of the \textit{BoWFire} method with the state-of-the-art methods.
	}
	\label{fig4:literature}
\end{figure*}

%\noindent
\textbf{Impact of $K_{sp}$:} Figure~\ref{fig4:superPixel} shows the results obtained while varying the number of superpixels.
Figure~\ref{fig4:superPixelFire} shows the results for the fire dataset.
In this case, there was a slight increase of all measure until $K_{sp}=150$, then for greater values they had a similar behavior.
The Precision obtained was around 0.8, Recall around 0.65 and F1-Score around 0.72.
Figure~\ref{fig4:superPixelNotFire} shows the results for the non-fire dataset.
For this dataset we computed only the False-Positive Rate.
There is a slight increasing of FPR as the number of superpixels increases, except for $K_{sp} = 300$.
It is important to notice that although FPR increases, the values remain around 0.045, that is, less than 5\% of false-positives.
And Figure~\ref{fig4:superPixelComplete} shows the results combining both datasets.
Again, there is a similar behavior regarding $K_{sp}$, except for  $K_{sp} = 50$.
The Precision obtained was around 0.5, Recall around 0.65 and F1-Score around 0.57.
The FPR values were not shown on both Figures~\ref{fig4:superPixelFire} and \ref{fig4:superPixelComplete} due to their low values for all $K_{sp}$.
There is also a slight increasing of FPR as $K_{sp}$ increases, but with lower values.
On the fire dataset, FPR went from 0.0169 to 0.0175, and on the complete dataset it varied from 0.0305 to 0.0323.

The main goal of the \textit{BoWFire} is to decrease the FPR while maintaining a good performance.
With that in mind, we evaluated that the best result is achieved when the number of superpixels $K_{sp} = 150$.
This number presented better results while dealing with just the fire and complete dataset (fire and non-fire), as showed by F1-score.
Also, the value of FPR for this $K_{sp}$ is close to the lowest FPR value.

%\noindent
\textbf{\textit{Color Classification} Evaluation:} Figures~\ref{fig4:colorClassFire}, \ref{fig4:colorClassNotFire} and \ref{fig4:colorClassComplete}, show the results for the \textit{Color Classification} step.
Considering only color-based approaches, \textit{Color Classification}, Celik and Rudz presented the best overall performance.
Although Chen obtained the highest value of Precision, its Recall got the lowest value.
%As seen in Figures~\ref{fig4:sample1} and \ref{fig4:sample2}, Chen missed too many true-positive pixels.
%And finally, Rossi has the lowest overall performance.
As seen in Figures~\ref{fig4:sample1} and \ref{fig4:sample2}, Chen missed too many true-positive pixels and Rossi has the lowest overall performance.
We observed that in outdoor emergency situations, fire was not in the cluster with the higher values of V, as shown in Figures~\ref{fig4:sample1} and \ref{fig4:sample2}.
From now on, we will focus our analysis only on the methods with the best overall performance.

%Figure~\ref{fig4:colorClassFire} shows the results for the fire dataset.
Regarding to Precision, Rudz achieved the best value,  0.84 on the fire dataset and 0.31 on the complete dataset, while \textit{Color Classification} and Celik had similar behavior with values around 0.62 on the fire dataset and 0.24 on the complete.
On the other hand, \textit{Color Classification} achieved the highest value of Recall, 0.77 on both fire and complete dataset, followed by Celik, 0.63 on both fire and complete, and Rudz, 0.41 on both fire and complete.
Analyzing the F1-Score, \textit{Color Classification} and Celik methods outperformed Rudz by at most 23.6\% on the fire dataset with values of 0.68 to \textit{Color Classification}, 0.63 to Celik and 0.55 to Rudz.
On the complete dataset, all methods achieved similar F1-Score with the value of 0.35.

On the fire dataset, \textit{Color Classification} and Celik achieved similar values of FPR (0.05 and 0.04) and Rudz method achieved 0.01 FPR.
On the non-fire dataset, \textit{Color Classification}, Celik and Rudz achieved respectively 0.21, 0.15 and 0.08.
And on the complete dataset, \textit{Color Classification}, Celik and Rudz methods achieved respectively 0.13, 0.10 and 0.05.
On all datasets, Rudz achieved the best FRP value, less than 9\% of the pixels was incorrectly classified.
However, while discarding more false-positives, Rudz also discarded true-positives, reducing its Recall capability.
Except for FPR, \textit{Color Classification} had a similar behavior of Celik, but had better values of Recall and F1-Score.
Therefore, the \textit{Color Classification} outperformed the other methods.

%\noindent
\textbf{\textit{BoWFire} Evaluation:} Figures~\ref{fig4:textureClassFire}, \ref{fig4:textureClassNotFire} and \ref{fig4:textureClassComplete} show the results when added texture information.
It is possible to note that there was an overall improvement for all methods.
Regarding Precision, with the exception of Rudz and Chen, all methods  had a considerable improvement.
\textit{Color Classification} and Celik had a Precision improvement of up to 1.30$\times$ on the fire dataset and 2.28$\times$ on the complete dataset.
Rossi had the greatest improvement, 4.43$\times$ on fire dataset and 5.65$\times$ on the complete dataset.
This high improvement was due to the fact that Rossi, on outdoor images, detected other regions than fire, as shown on Figures~\ref{fig4:sample1} and \ref{fig4:sample2}.
When adding texture information, these false-positive regions were discarded.
For both Chen and Rudz, there was a slightly improvement on the fire dataset, but it is due to the fact that they already had low FPR.
On the other hand, on the complete dataset, there was an improvement of 1.64$\times$ to Chen and 2.06$\times$ to Rudz.

There was a decreasing on the Recall value of up to 15\% less for all methods, except Rudz.
This is due to the fact that the combination of both approaches discarded a few true-positives.
However, the considerable gain on the precision can justify this drawback.
%Analyzing the F1-Score, there was a slight increase of up to 7\% for all methods, except Rossi, on the fire dataset.
%Rossi had an improvement of 69\%.
Analyzing the F1-Score, there was a slight increase of up to 7\% for all methods, except Rossi, on the fire dataset, which had an improvement of 69\%.
On the complete dataset, all methods had a considerable improvement, up to 65\%.

As one of the goals of \textit{BoWFire} is to reduce the number of false-positives, it is important to analyze FPR.
On the fire dataset, there was a reduction of up to 68\% of FPR for \textit{Color Classification}, using Celik and Chen methods.
Rossi had 94\% less false-positives.
Rudz was the least affected by this step, reducing 5\% false-positives, since they had already dismissed false-positives on a post processing step.
On both the non-fire and complete dataset, all methods reduced FPR by up to 80\%.
This result confirms that the \textit{Texture Classification} step is capable of discarding false-positives without compromising the overall performance.

\begin{figure}[h!]
	\includegraphics[width=0.9\columnwidth]{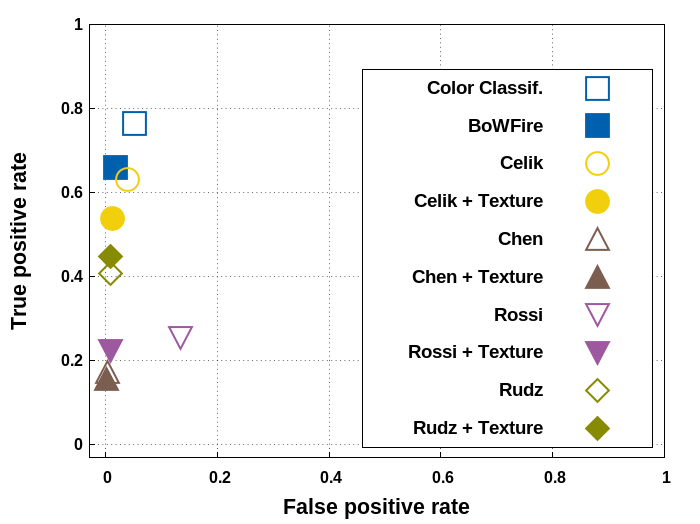}
	\caption{
		ROC Space using the fire dataset.
	}
	\label{fig4:rocFire}
\end{figure}

\begin{figure}[ht]
	\includegraphics[width=0.9\columnwidth]{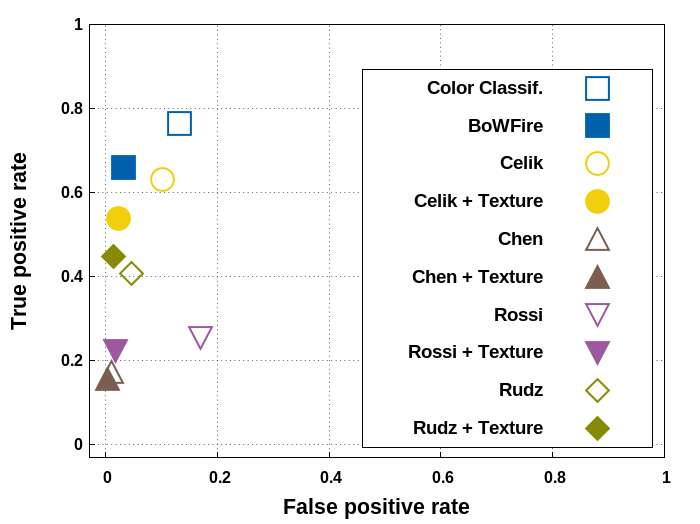}
	\caption{
		ROC Space using the complete dataset.
	}
	\label{fig4:rocComplete}
\end{figure}

\begin{figure}[ht]
	\centering
\begin{subfigure}[b]{0.20\textwidth}
	\includegraphics[width=\textwidth]{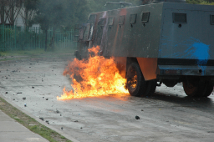}
	\caption{Input image}
\end{subfigure}
\begin{subfigure}[b]{0.20\textwidth}
	\includegraphics[width=\textwidth]{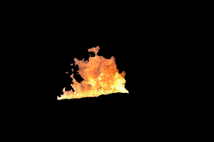}
	\caption{Ground truth}
\end{subfigure}
\begin{subfigure}[b]{0.20\textwidth}
	\includegraphics[width=\textwidth]{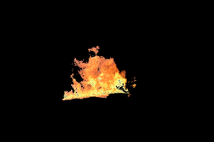}
	\caption{\textit{BoWFire}}
\end{subfigure}
\begin{subfigure}[b]{0.20\textwidth}
	\includegraphics[width=\textwidth]{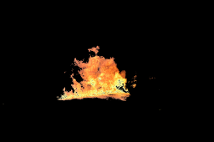}
	\caption{\textit{Color Classification}}
\end{subfigure}
\begin{subfigure}[b]{0.20\textwidth}
	\includegraphics[width=\textwidth]{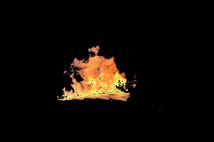}
	\caption{Celik}
\end{subfigure}
\begin{subfigure}[b]{0.20\textwidth}
	\includegraphics[width=\textwidth]{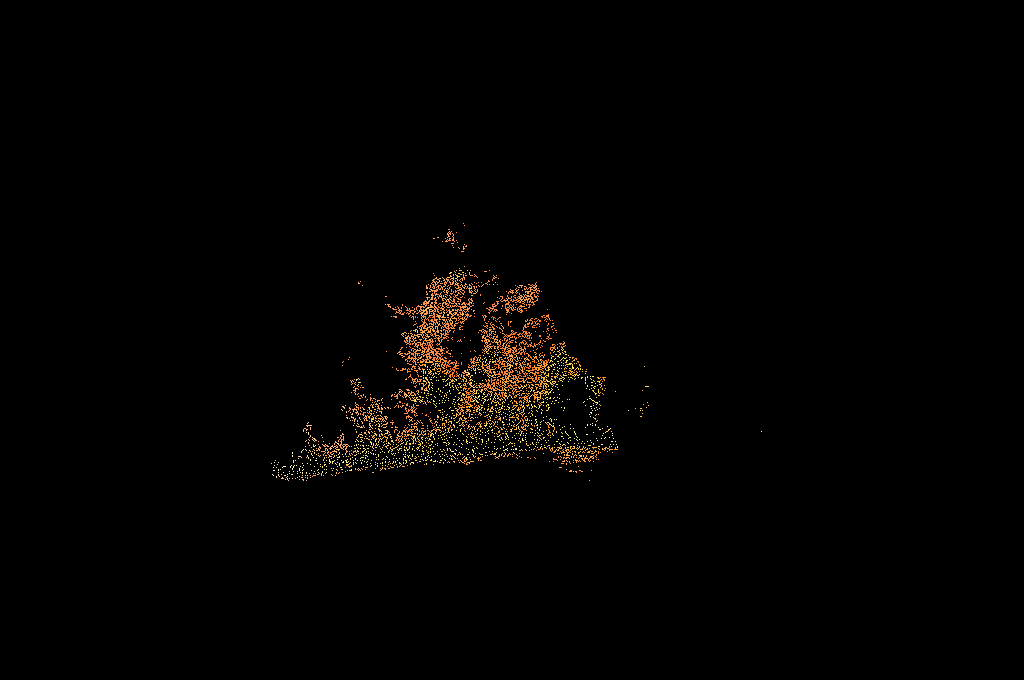}
	\caption{Chen}
\end{subfigure}
\begin{subfigure}[b]{0.20\textwidth}
	\includegraphics[width=\textwidth]{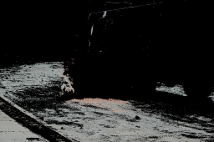}
	\caption{Rossi}
\end{subfigure}
\begin{subfigure}[b]{0.20\textwidth}
	\includegraphics[width=\textwidth]{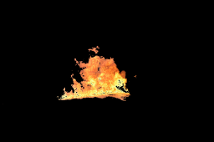}
	\caption{Rudz}
\end{subfigure}

	\caption{
		Output from the methods with an input image with fire.
	}
	\label{fig4:sample1}
\end{figure}

\begin{figure}[ht]
	\centering
\begin{subfigure}[b]{0.20\textwidth}
	\includegraphics[width=\textwidth]{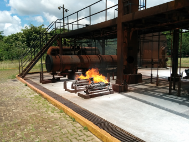}
	\caption{Input image}
\end{subfigure}
\begin{subfigure}[b]{0.20\textwidth}
	\includegraphics[width=\textwidth]{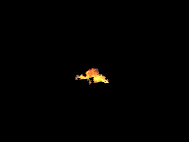}
	\caption{Ground truth}
\end{subfigure}
\begin{subfigure}[b]{0.20\textwidth}
	\includegraphics[width=\textwidth]{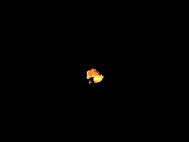}
	\caption{\textit{BoWFire}}
\end{subfigure}
\begin{subfigure}[b]{0.20\textwidth}
	\includegraphics[width=\textwidth]{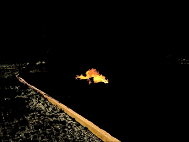}
	\caption{\textit{Color Classification}}
\end{subfigure}
\begin{subfigure}[b]{0.20\textwidth}
	\includegraphics[width=\textwidth]{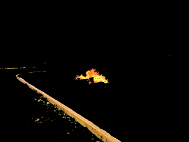}
	\caption{Celik}
\end{subfigure}
\begin{subfigure}[b]{0.20\textwidth}
	\includegraphics[width=\textwidth]{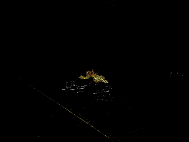}
	\caption{Chen}
\end{subfigure}
\begin{subfigure}[b]{0.20\textwidth}
	\includegraphics[width=\textwidth]{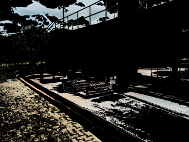}
	\caption{Rossi}
\end{subfigure}
\begin{subfigure}[b]{0.20\textwidth}
	\includegraphics[width=\textwidth]{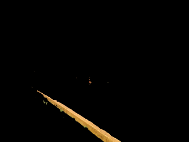}
	\caption{Rudz}
\end{subfigure}

	\caption{
		Output from the methods with an input image with fire and possible false-positives pixels.
	}
	\label{fig4:sample2}
\end{figure}

%\sout{The previous results demonstrated that adding texture information improves the overall performance.}
We can now use the Receiver Operating Characteristic (ROC) space to analyze the performance behavior between all methods.
The ROC Space shows the relation between FPR and the true-positive rate (Recall). %\sout{, where FPR is on the ordinate axis and Recall is on the abscissa axis.}
%\sout{On the ROC Space the nearest the point to $(0, 1)$, the better the classifier.}
Figures~\ref{fig4:rocFire} and \ref{fig4:rocComplete} show the ROC Space on, respectively, the fire and the complete datasets for all methods.
On both ROC Spaces, it is possible to note that all methods move to the left, i.e., achieve less FPR when texture information is added.
The \textit{Color Classification} and the \textit{BoWFire} method presented the best classification results among the other methods, followed by Celik.
Also, the \textit{BoWFire} achieved a similar Recall value as Celik without texture information, but with a smaller FPR.

\begin{figure}[ht]
	\centering
\begin{subfigure}[b]{0.20\textwidth}
	\includegraphics[width=\textwidth]{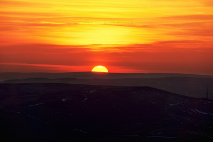}
	\caption{Input image}
\end{subfigure}
\begin{subfigure}[b]{0.20\textwidth}
	\includegraphics[width=\textwidth]{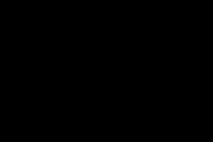}
	\caption{\textit{BoWFire}}
\end{subfigure}
\begin{subfigure}[b]{0.20\textwidth}
	\includegraphics[width=\textwidth]{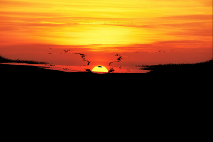}
	\caption{\textit{Color Classification}}
\end{subfigure}
\begin{subfigure}[b]{0.20\textwidth}
	\includegraphics[width=\textwidth]{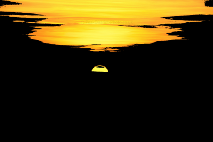}
	\caption{Rudz}
\end{subfigure}

	\caption{
		Output of a non-fire image.
	}
	\label{fig4:sample3}
\end{figure}

Figures~\ref{fig4:sample1}, \ref{fig4:sample2} and \ref{fig4:sample3} show visual samples of output images from three different situations.
Figure~\ref{fig4:sample1} shows an emergency situation with fire and low percentage of possible false-positives.
On this input image it is possible to note that \textit{Color Classification}, Celik and Rudz methods had similar outputs.
The \textit{BoWFire} method was able to detect the same fire region as these methods but discarded the fire reflection on the ground.
Rossi was not able to correctly detect fire regions, while Chen discarded more than half of the true-positives.
Figure~\ref{fig4:sample2} also shows an emergency situation with fire with a higher percentage of false-positives.
In this case, all methods detected false-positives, with the exception of \textit{BoWFire}.
It is also possible to note that in some cases, Rudz discards more fire pixels than necessary.
This image also shows the problem with the Rossi method, since no fire region was detected as fire.
Once again, Chen discarded almost every true-positive.
Figure~\ref{fig4:sample3} shows a sunset skyline image.
For this input image, both \textit{Color Classification} and Rudz detected false-positives. %, with FPR of 0.47 and 0.24 respectively.
Meanwhile, when adding texture information to the \textit{Color Classification}, \textit{BoWFire} was capable of discarding all false-positives for this image.

\section{Conclusions}
\label{sec:conclusion}

In this paper, we presented the \textit{BoWFire} method, a novel approach for fire detection on images in emergency context.
Our results showed that \textit{BoWFire} was capable of detecting fire with a perfomance similar to what is observed in the works of the state-of-the-art, but with less false-positives.
We systematically compared our work with four former works, demonstrating that we achieved consistent improvements. 

The course of action of \textit{BoWFire} was that, by simultaneously using color and texture information, it was able to dismiss false-positives relying solely on the information present in the images; as opposed to former methods that use temporal information.
Furthermore, since \textit{BoWFire} is based on classification methods, rather than on mathematical modeling, it was able to solve the problem with only three parameters.
In addition, these parameters were more intuitive for tuning, unlike those of previous works, which are based on thresholds and color-based values.
Given its performance, we conclude that BoWFire is suitable to integrate a crisis management system as the one that motivates this work. 

%More important, the \textit{BoWFire} was able to dismiss false-positives with just the information present in the images.
%Another positive feature of the proposed method was the small number of parameters needed for its execution.
%Also, the parameters of the \textit{BoWFire} were more intuitive for tuning, unlike those of previous works, which need thresholds and color-based values.
%Given its performance, we conclude that BoWFire is suitable to integrate a crisis management system. 

%Rescue specialists need the most relevant data available to make fast and accurate decisions.
%Volume, redundancy and presence of non significant content on image data may be harmful to specialists handling with emergency situations.
%With the advantages stated earlier, we conclude that BoWFire is suitable to be integrated in a crisis management system because of its overall performance in detecting fire incidents related to emergency events from still images data.

As future work, we plan to investigate the combination of different features extraction methods.
%Also, we intend to explore more refined classification algorithm for both the \textit{Color} and \textit{Texture Classification} steps.
We also envision the extension of \textit{BoWFire} to detect other types of incident, such as smoke and explosions.

%The \textcolor{red}{RESCUER} Project provides a crowdsource solution to automatically provide data analysis for a Crisis Control Committee (CCC).
%This way, the victims and rescue team present at the emergency location can send images and videos of the incident.
%With the images of the emergency situation, the CCC can aid the local rescue team by providing extra information regarding the type of incident and its current status.
%Although, the volume and redundancy of the image data may dispend a huge amount of resource of the CCC to manually analyze.
%Thus, it is important to discard unnecessary data and summarize the relevant information.

\iffinal
\section*{Acknowledgment}
%
%The authors would like to thank this colleague and this financing institute.
This research is partially funded by FAPESP, CNPq, CAPES, STIC-AmSud and \textit{RESCUER} Project, funded by the European Commission (Grant: 614154) and by the CNPq/MCTI (Grant: 490084/2013-3).
\fi

%==========================================


\bibliographystyle{IEEEtran}

\end{document}